# Statistical Vs Rule Based Machine Translation;
# A Case Study on Indian Language Perspective

Sreelekha S.

Dept. of Computer Science & Engineering, Indian Institute of Technology Bombay, India
sreelekha@cse.iitb.ac.in

**Abstract.** In this paper we present our work on a case study between Statistical Machien Transaltion (SMT) and Rule-Based Machine Translation (RBMT) systems on English-Indian langugae and Indian to Indian langugae perspective. Main objective of our study is to make a five way performance compariosn; such as, a) SMT and RBMT b) SMT on English-Indian langugae c) RBMT on English-Indian langugae d) SMT on Indian to Indian langugae perspective e) RBMT on Indian to Indian langugae perspective. Through a detailed analysis we describe the Rule Based and the Statistical Machine Translation system developments and its evaluations. Through a detailed error analysis, we point out the relative strengths and weaknesses of both systems. The observations based on our study are: a) SMT systems outperforms RBMT b) In the case of SMT, English to Indian language MT systmes performs better than Indian to English langugae MT systems c) In the case of RBMT, English to Indian langugae MT systems perofrms better than Indian to Englsih Language MT systems d) SMT systems performs better for Indian to Indian language MT systems compared to RBMT. Effectively, we shall see that even with a small amount of training corpus a statistical machine translation system has many advantages for high quality domain specific machine translation over that of a rule-based counterpart.

## 1  Introduction

Machine Translation (MT) is an area of research that combines ideas and techniques from Linguistics, Computer Science, Artificial Intelligence, Translation theory and Statistics for automating the process of translation from one language to another. Major difficulties in MT are the difference between the source and target languages and their ambiguities.

There are many ongoing attempts to develop MT systems for regional languages using various approaches [2]. The approaches to machine translation are categorized as, Rule Based or Knowledge Driven approaches and Corpus Based or Data-Driven approaches. The RBMT approaches are further classified into Transfer based MT, Interlingua MT and Dictionary based MT, while the Corpus Based approaches are classified into Example Based MT and SMT. Many studies have been conducted in the case of English to Indian languages and Indian to Indian languages MT system development [3], [15], [16], [17], [18], [19], [20]. This paper discusses a comparative study on RBMT and SMT approaches used in English to Indian languages and Indian to Indian language MT systems.

The organization of the paper is as follows: Section 2 starts with the discussion about rule based and statistical based MT approaches; Section 3 presents the Experiments conducted, Evaluations and Error analysis which convey the main components of the paper; Section 4 concludes the paper.
.

## 2  RULE BASED VS. STATISTICAL

RBMT system requires a huge human effort to prepare the rules and linguistic resources, such as morphological analyzers, part-of-speech taggers and syntactic parsers, bilingual dictionaries, transfer rules, morphological generator and reordering rules etc. In the case of English to Indian languages and Indian to Indian languages, there have been fruitful attempts with all the four RBMT approaches [5], [15], [12], [16], [17], [18], [19], [20]. Data-driven approaches, which provides an alternative to direct and rule-based MT systems have come to the fore of language processing research over the past decade. These approaches use a supervised or unsupervised statistical machine learning algorithm to build statistical models from the bilingual parallel corpora. There are three different statistical approaches in MT, Word-based Translation, Phrase-based Translation, and Hierarchical phrase based model. This paper discusses phrase based statistical approaches used against Rule based approaches in English-Indian language and Indian-Indian language MT systems to generate quality translations.

### 2.1 Rule Based Machine Translation

Rule based MT systems works based upon the specification of rules for morphology, syntax, lexical selection and transfer and generation. Collection of rules and a bilingual or multilingual lexicon are the resources used in RBMT. In the case of English to Indian languages and Indian language to Indian language MT systems, there have been many attempts with all these approaches [20]. The transfer model involves three stages: analysis, transfer and generation. Figure 1 shows the complete work flow of translation in the form of a pipeline.

During analysis phase linguistic analysis is performed on the input source sentence in order to extract information in terms of morphology, parts of speech, phrases, named entity and word sense disambiguation. During the lexical transfer phase, there are two steps namely word translation and grammar translation. In word translation, source language root word is replaced by the target language root word with the help of a bilingual dictionary and in grammar translation, suffixes are getting translated. In generation phase genders of the translated words are corrected and it will be followed by short distance and long-distance agreements performed by intra-chunk and the inter-chunk module. These ensure that the gender, number and person of local groups of phrases agree as also the gender of the subject's verbs or objects reflect those of the subject.

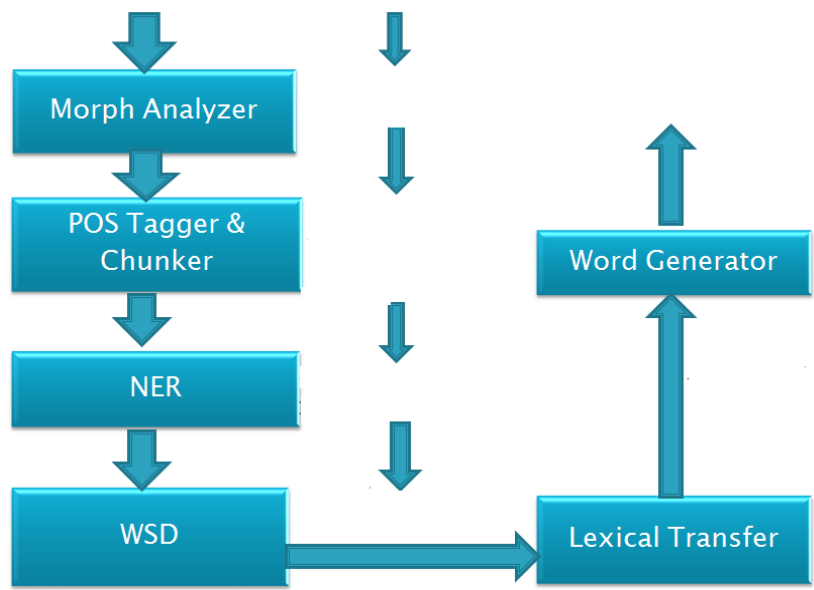

**Fig. 1.** RBMT work flow

### 2.2 Statistical Machine Translation

The statistical approach works based up on the statistical models extracted from parallel aligned bilingual text corpora, which takes the assumption that every word in the target language is a translation of the source language words with some probability [7], [8], [13]. The words which have the highest probability will give the best translation. Consistent patterns of divergence between the languages [2], [6], [20] when translating from one language to another, handling reordering divergence are one of the fundamental problems in MT. Figure 2 shows the functional flow diagram of an SMT system. The major steps in SMT are: Corpus preparation, Training, Decoding and Testing.

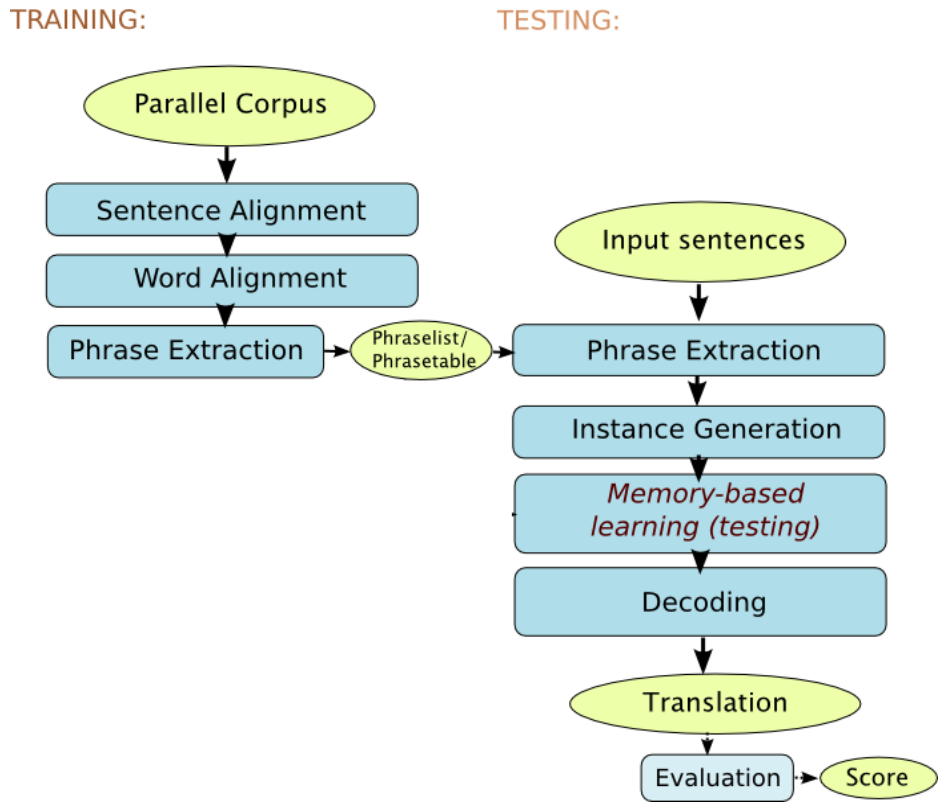

**Fig. 2.** SMT work flow

Corpus preparation, alignment and its cleaning will be done in the Pre-Processing step. Training is a process in which a supervised or unsupervised statistical machine learning algorithm is used to build statistical tables from the parallel corpora [13]. In Statistical Machine Translation, word by word and phrase based alignment plays the major role during parallel corpus training. During training Translational model, Language Model, Distortion Table, Phrase table etcetera are modelled. Decoding [7], [8], [10] is the most complex task in Machine Translation [10] where the trained models will be decoded. It is the major process in which the target language translations are being decoded using the generated phrase table, translation model and language model. The two major concerns with SMT are decoding complexity and target language reordering [1].

## 3     EXPERIMENTAL DISCUSSIONS

| Sl. No. | Corpus Source | Training Corpus [Manually cleaned and aligned] | Corpus Size [Sentences] |
|---|---|---|---|
| 1 | ILCI | Tourism | 23500 |
| 2 | ILCI | Health | 23500 |
| | | Total | 47000 |
| Sl. No. | Corpus Source | Tuning Corpus [Manually cleaned and aligned] | Corpus Size [Sentences] |
| 1 | ILCI | Tourism | 250 |
| 2 | ILCI | Health | 250 |
| | | Total | 500 |
| Sl. No. | Corpus Source | Testing Corpus [Manually cleaned and aligned] | Corpus Size [Sentences] |
| 1 | ILCI | Tourism | 1000 |

| Sl. No. | Corpus Source | Testing Corpus (Subjective Evaluation) [Manually cleaned and aligned] | Corpus Size [Sentences] |
|---|---|---|---|
| 2 | ILCI | Health | 1000 |
| | | Total | 2000 |
| 1 | ILCI | Tourism | 250 |
| 2 | ILCI | Health | 250 |
| | | Total | 500 |

**Table 1.** Corpus Statistics

We now describe the SMT system experiments performed and the comparisons with the results, in the form of an error analysis, of the Rule Based system described above. For the purpose of constructing with statistical models we use Moses [14] and Giza++.

Our experiments are focused on two research directions;
1) Indian- Indian language Perspective[1,2]
2) English Indian Language perspective[3,4]

For Indian-Indian language MT System case study we have used Marathi-Hindi as base language pairs and for English – Indian Language MT system case study we have used English- Malayalam as base language pairs.

### 3.1 Statistical Machine Translation System Experiments

We manually cleaned a 90000 sentence parallel corpus for both Marathi- Hindi and English Malayalam language pairs. We have corrected the grammatical structure of the sentences and tokenized it thereby making available a high quality corpus for training. Table 1 describes the corpus resources we have used for training. We followed the training steps of moses baseline system. In order to perform tuning we used 500 sentence pairs. We observed that there was only slight improvement on the translation quality since the sentence pairs used for tuning had a number of stylistic constructions and bleu based tuning tends to cause deterioration of quality. We have tested the translation system with a corpus of 1000 sentences taken from the 'ILCI tourism health' corpus as shown in table 1. The added advantage in the case of Marathi-Hindi compared to English- Malayalam was the SOV ordering similarity between Marathi and Hindi. However there were difficulties in handling inflected words.

**Evaluation.**
To analyze the quality of translation, we have used both subjective evaluation and bleu score [11] evaluation. In order to evaluate the correct grammatical constructions present in the translated sentence, we have used Fluency as an indicator whereas the amount of meaning being carried over from the source to the target is indicated by adequacy measure. Depending on how much sense the translation made and its grammatical correctness, we assigned scores between 1 and 5 for each translation. The basis of scoring is given below:
- 5: If the translations are perfect.
- 4: If there are one or two incorrect translations and mistakes.
- 3: If the translations are of average quality, barely making sense.
- 2: If the sentence is barely translated.
- 1: If the sentence is not translated or the translation is gibberish.

S1, S2, S3, S4 and S5 are the counts of the number of sentences with scores from 1 to 5 and N is the total number of sentences evaluated. The formula [9] used for computing the scores is:

$$A/F = 100 * ((S5 + 0.8 * S4 + 0.6 * S3))/N$$

---

[1] http://tdil-dc.in/mt/common.php#
[2] http://www.cfilt.iitb.ac.in/SMT-System/
[3] http://www.cfilt.iitb.ac.in/SMT-EM/
[4] http://www.cfilt.iitb.ac.in/SMT-ME/

We considered the sentences with scores above 3 only and we penalize the sentences with scores 4 and 3 by multiplying their count by 0.8 and 0.6 respectively in order to make the estimate of scores is much better. The results of our evaluations are given below in Table 4.

### 3.2 English-Indian Language case study results

In order to perform English-Indian language case study, we have used English-Malayalam and Malayalam-English as base language pairs. The results of bleu score evaluation and subjective evaluations are shown in table 2, 3, 4 and 5.

| English- Malayalam MT System | Adequacy | Fluency |
|---|---|---|
| Rule Based | 55.6% | 47% |
| Statistical | 77.23% | 87% |

Table 2. Results of English- Malayalam SMT Vs. RBMT Subjective Evaluation

| English-Malayalam MT System | BLEU Score |
|---|---|
| Rule Based | 20.8 |
| Statistical | 39.90 |

Table 3. : Results of English- Malayalam SMT Vs. RBMT BLEU score

| Malayalam- English MT System | Adequacy | Fluency |
|---|---|---|
| Rule Based | 64.6% | 51% |
| Statistical | 74.89% | 85.34% |

Table 4. Results of Malayalam- English SMT Vs. RBMT Subjective Evaluation

| Malayalam- English MT System | BLEU Score |
|---|---|
| Rule Based | 29.9 |
| Statistical | 37.90 |

Table 5. : Results of Malayalam- English SMT Vs. RBMT BLEU score

| Marathi- Hindi MT System | Adequacy | Fluency |
|---|---|---|
| Rule Based | 69.6% | 58% |
| Statistical | 79.8% | 88.4% |

Table 6. Results of Marathi-Hindi SMT Vs. RBMT Subjective Evaluation

| Marathi-Hindi MT System | BLEU Score |
|---|---|
| Rule Based | 23.3 |
| Statistical | 51.60 |

Table 7. : Results of Marathi-Hindi SMT Vs. RBMT BLEU score

| Hindi- Marathi MT System | Adequacy | Fluency |
|---|---|---|
| Rule Based | 64.8% | 56.78% |
| Statistical | 75.89% | 85.14% |

Table 8. Results of Hindi- Marathi SMT Vs. RBMT Subjective Evaluation

| Hindi- Marathi MT System | BLEU Score |
|---|---|
| Rule Based | 17.9 |
| Statistical | 43.30 |

Table 9. : Results of Hindi- Marathi SMT Vs. RBMT BLEU score

### 3.3 Indian to Indian language case study results

In order to perform Indian-Indian language MT case study, we have used Marathi-Hindi and Hindi-Marathi system as base pairs. The results of bleu score evaluation and subjective evaluations are shown in Table 6, 7, 8 and 9.

### 3.4 SMT Vs RBMT Analysis

Table 10. Comparison of SMT over RBMT

| Sl. No | Performance comparison of SMT over RBMT | |
|---|---|---|
| | SMT | RBMT |
| 1 | Being able to handle rich morphology, can easily separate suffixes from inflected words with gender number person aspect and mood, leading to meaning transfer | Not able to split suffixes from inflected words with gender number person aspect and mood by itself and hence fails to handle rich morphology |
| 2 | Able to handle verb phrases and function words since SMT follows memory based training to learn phrase | Unable to effectively handle the appropriate translation and generation of function words, verb phrases etc |
| 3 | Rapid, easier to create, maintain and improve upon, in short cost-effective development | Robust, high development and customization cost |
| 4 | Can handle ambiguity since it records phrase translations with its frequency of occurrence which acts as more natural word sense disambiguation | Fails to handle ambiguity due to poor quality WSD approaches |
| 5 | Good fluency and adequacy due to plentiful evidences of good quality phrase pairs recorded in phrase table | Lack of fluency |
| 6 | Language model used helped in generating more natural translations | Morph analyzers process word by word and hence fails to generate natural translations |
| 7 | Data driven hence domain specific | Knowledge driven hence can work for out of domain data also |

Table 101. Performance comparison of English – Malayalam SMT over Malayalam - English SMT

| | Performance comparison of English – Malayalam SMT over Malayalam - English SMT | |
|---|---|---|
| | English – Malayalam SMT | Malayalam - English SMT |
| 1 | Malayalam agglutinative suffixes have English equivalents in the form of pre positions. So during alignment from English to Malayalam, English word can align to the words with agglutination in Malayalam, since it is a single word. | While aligning from Malayalam - English the agglutinative word may map only to root words, there is a chance to miss out the pre position mapping in English, since it is separate words. |
| 2 | Require Morphology Generation for Malayalam | Require Morphology Analysis for Malayalam |
| 3 | Rapid, easier to create, maintain and improve upon, in short cost-effective development | Rapid, easier to create, maintain and improve upon, in short cost-effective development |
| 4 | Good fluency and adequacy, since there is more probability to get map to the inflected Malayalam word from English word. | Less fluency, since multiple words have to get mapped from a single inflected form during translation is more erroneous. |
| 5 | Language model used helped in generating more natural translations | Language model used helped in generating more natural translations |

**Table 11**2. Performance comparison of Malayalam - English RBMT over English – Malayalam RBMT

| Sl. No | Performance comparison of Marathi - Hindi SMT over Hindi – Marathi SMT | |
|---|---|---|
| | Marathi - Hindi SMT | Hindi – Marathi SMT |
| 1 | Marathi agglutinative suffixes have Hindi equivalents in the form of post positions. Since Marathi and Hindi have same SOV order it can easily map the inflections to a great level | While aligning form Hindi to Marathi, there is a probability that the agglutinative words may miss out from the post position mapping from Hindi, since it is separate words in many cases compared to Marathi. |
| 2 | Require Morphology Analysis for Marathi | Require Morphology Generation for Marathi |
| 3 | Rapid, easier to create, maintain and improve upon, in short cost-effective development | Rapid, easier to create, maintain and improve upon, in short cost-effective development |
| 4 | Good fluency and adequacy, since it is easy to map from the Marathi word to the Hindi equivalent form. | Less fluency, since a single inflected word has to map from multiple words during translation is more erroneous. |
| 5 | Language model used helped in generating more natural translations | Language model used helped in generating more natural translations |

**Table 12**3. Performance comparison of Marathi - Hindi SMT over Hindi – Marathi SMT

| Sl. No | Performance comparison of Malayalam - English RBMT over English – Malayalam RBMT | |
|---|---|---|
| | Malayalam - English RBMT | English – Malayalam RBMT |
| 1 | Agglutinated Malayalam suffixes have English equivalents in the form of pre positions. | While translating from English - Malayalam word by word processing of analysis and generation may not help to generate the correct agglutinative Malayalam word formation. |
| 2 | Require Morphology Analysis for Malayalam | Require Morphology Generation for Malayalam |
| 3 | During Morphology Analysis from a single inflected word, agglutinated suffixes are getting separated and equivalent group words are translated during lexical transfer. | During Morphology generation from a group of English words, all words may not get properly formed. There is higher chance to get error in proper generation of inflected form. |
| 4 | Generating pre positioned English words are easy | Generating rich morphological Malayalam agglutinative suffixed words are difficult. |
| 5 | Fluency and adequacy will be more | Fluency and adequacy will be less |

We have done a detailed error analysis on both RBMT and SMT systems. Table 10 shows the observations during the case study analysis. Further we explain the observations of a detailed case study between English-Malayalam and Marathi-Hindi language pairs with SMT and RBMT experiments. Table 11, 12, 13 and 14 shows the Performance comparison analysis results of various aspects of SMT and RBMT approaches over Indian-Indian language and English-Indian language MT systems.

**Table 13**4. Performance comparison of Marathi - Hindi RBMT over Hindi – Marathi RBMT

| Sl. No. | Performance comparison of Marathi - Hindi RBMT over Hindi – Marathi RBMT | |
|---|---|---|
| | Marathi - Hindi RBMT | Hindi – Marathi RBMT |
| 1 | Marathi suffixes have Hindi equivalents in the form of post positions. So during analysis from Marathi to Hindi, group of Hindi words have to generate. | From group of Hindi words agglutinative Marathi inflected form have to generate. |
| 2 | Require Morphology Analysis for Marathi and generator for Hindi | Require Morphology Analysis for Hindi and Morphology Generation for Marathi |
| 3 | During Morphology Analysis from a single inflected word, agglutinated suffixes are getting separated and equivalent Hindi words are translated during lexical transfer. | During Morphology generation from word by word, all words may not get properly formed to generate the correct Marathi word. There is higher chance to get error in proper generation of inflected form. |
| 4 | Generating post positioned Hindi words are easy | Generating rich morphological Marathi agglutinative suffixed words are difficult. Morph analyzers process word by word and hence fails to generate natural Marathi translations |
| 5 | Fluency and adequacy will be more | Fluency and adequacy will be less |

Figure 3 and 4 shows SMT Vs RBMT evaluation graphs for the English-Indian and Indian-Indian Language case study reults.

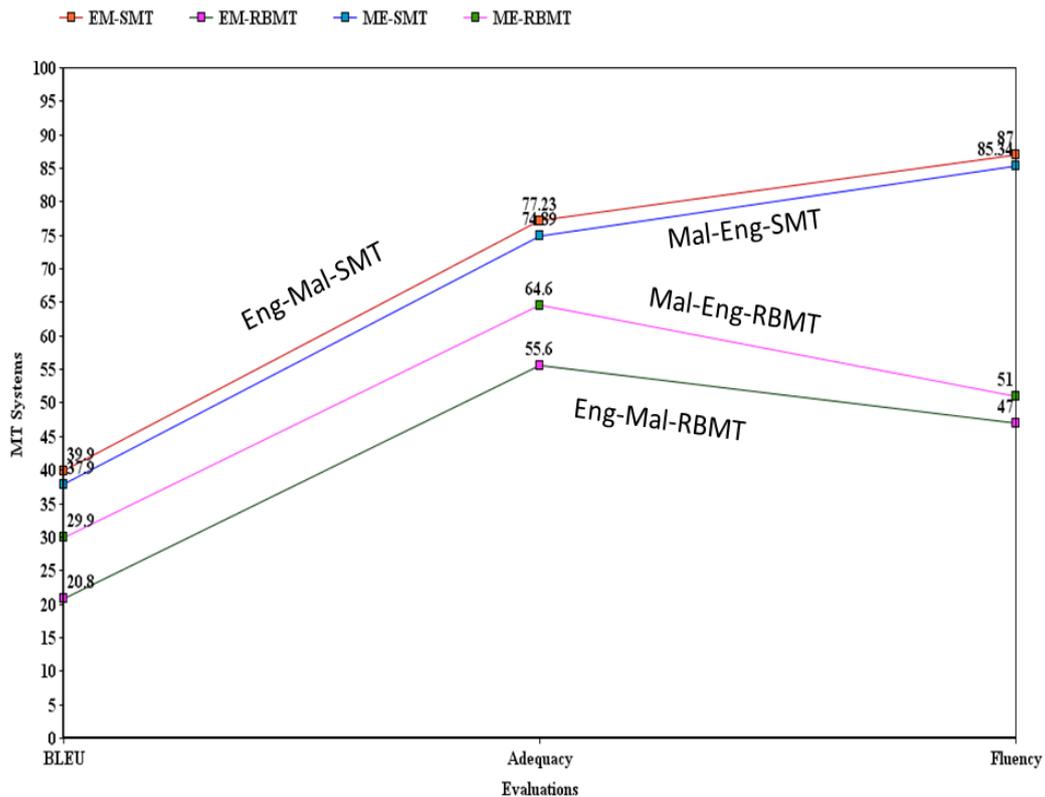

**Fig. 3.** SMT Vs RBMT English-Indian Language Evaluations graph

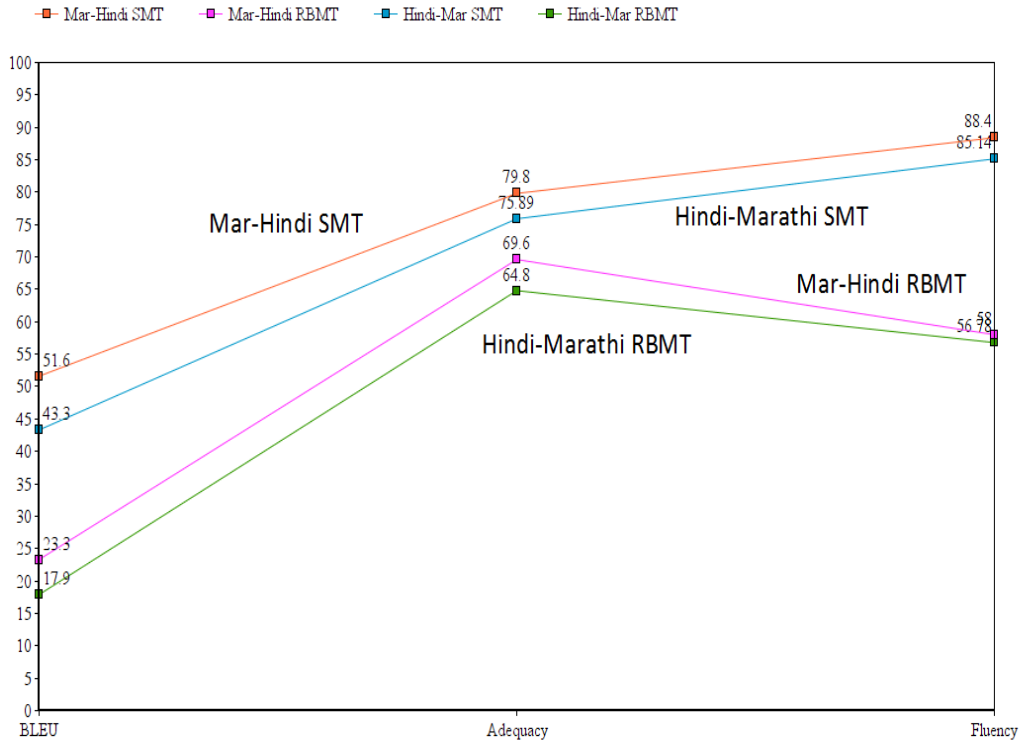

**Fig. 4.** SMT Vs. RBMT Indian-Indian Language Evaluations graph

## 4  CONCLUSIONS

In this paper we have mainly focused on the comparative performance of Statistical Machine Translation and Rule- Based Machine Translation on Indian to Indian language perspective and English to Indian language perspective.  Our major observations are,
1. Translation quality of SMT is relatively high as compared to the RBMT system, considering that the efforts required to build RBMT systems is huge.
2. SMT perform better for English to Malayalam systems comparing to Malayalam to English systems.
3. RBMT performs better for Malayalam to English as of English to Malayalam
4. SMT system performs better for Marathi –Hindi compared to Hindi-Marathi
5. RBMT performs better for Marathi- Hindi compared to Hindi-Marathi
6. For English-Indian language scenario, SMT performs better for morphologically low language to rich language and on the other hands RBMT performs better for morphologically rich language to low language.
7. Indian to Indian language MT performs better than English to Indian language MT in terms of SMT.
8. English to Indian language MT performs better than Indian to Indian language MT in terms of RBMT.

We observed that translation quality of Statistical Machine Translation is relatively high than the Rule Based system, since the efforts required to build RBMT systems is huge. Also SMT which cannot split suffixes by itself was unable to handle the translation of inflected suffix words in some cases. RBMT being able to use the morph analyzer can easily separate the suffixes from the inflected words and generate translations.

## ACKNOWLEDGEMENT

The authors would like to thank Department of Science & Technology, Govt. of India for providing fund under Woman Scientist Scheme (WOS-A) with the project code-SR/WOS-A/ET/1075/2014.